\documentclass[runningheads]{llncs}
\usepackage{graphicx}

\usepackage{tikz}
\usepackage{comment}
\usepackage{amsmath,amssymb} 
\usepackage{color}

\usepackage{graphicx}
\usepackage{amsmath}
\usepackage{amssymb}
\usepackage{booktabs}

\usepackage{epsfig}
\usepackage{tabularx}
\usepackage{multirow}
\usepackage{subfloat}
\usepackage{caption}
\usepackage{comment}
\usepackage{makecell}

\usepackage{xcolor}
\definecolor{citecolor}{HTML}{0071bc}
\usepackage[pagebackref,breaklinks,colorlinks,citecolor=citecolor,bookmarks=false]{hyperref}

\usepackage[capitalize]{cleveref}
\crefname{section}{Sec.}{Secs.}
\Crefname{section}{Section}{Sections}
\Crefname{table}{Table}{Tables}
\crefname{table}{Tab.}{Tabs.}
\usepackage[accsupp]{axessibility}  

\begin{document}
\pagestyle{headings}
\mainmatter
\def\ECCVSubNumber{4675}  

\title{Learning Implicit Feature Alignment Function for  Semantic Segmentation} 

\titlerunning{Implicit Feature Alignment Function}
%
\author{Hanzhe Hu\inst{1}* \and
Yinbo Chen\inst{2}*  \and 
Jiarui Xu\inst{2}  \and 
Shubhankar Borse \inst{3}  \and 
Hong Cai \inst{3} \and
Fatih Porikli \inst{3} \and
Xiaolong Wang \inst{2}}
\authorrunning{H. Hu et al.}
%
\institute{Peking University \and
UC San Diego \and
Qualcomm AI Research\\
}

\maketitle

\begin{abstract}
Integrating high-level context information with low-level details is of central importance in semantic segmentation. Towards this end, most existing segmentation models apply bilinear up-sampling and convolutions to feature maps of different scales, and then align them at the same resolution. However, bilinear up-sampling blurs the precise information learned in these feature maps and convolutions incur extra computation costs. To address these issues, we propose the Implicit Feature Alignment function (IFA). Our method is inspired by the rapidly expanding topic of implicit neural representations, where coordinate-based neural networks are used to designate fields of signals. In IFA, feature vectors are viewed as representing a 2D field of information. Given a query coordinate, nearby feature vectors with their relative coordinates are taken from the multi-level feature maps and then fed into an MLP to generate the corresponding output. As such, IFA implicitly aligns the feature maps at different levels and is capable of producing segmentation maps in arbitrary resolutions. We demonstrate the efficacy of IFA on multiple datasets, including Cityscapes, PASCAL Context, and ADE20K. Our method can be combined with improvement on various architectures, and it achieves state-of-the-art computation-accuracy trade-off on common benchmarks. Code will be made available at \href{https://github.com/hzhupku/IFA}{https://github.com/hzhupku/IFA}.
\end{abstract}


\section{Introduction}

\renewcommand{\thefootnote}{\fnsymbol{footnote}}
\footnotetext[0]{*\ Equal contribution.}

Semantic Segmentation is one of the most fundamental and challenging tasks in computer vision. It aims at classifying each pixel in the image into a semantic category. Its wide applications include scene understanding, image editing, augmented reality, and autonomous driving. Most of these applications not only require the segmentation model to predict high resolution and high-quality masks but also demand high efficiency in speed and memory cost, especially when running online or on edge devices.

\begin{figure}
    \centering
    \includegraphics[width=0.75\linewidth]{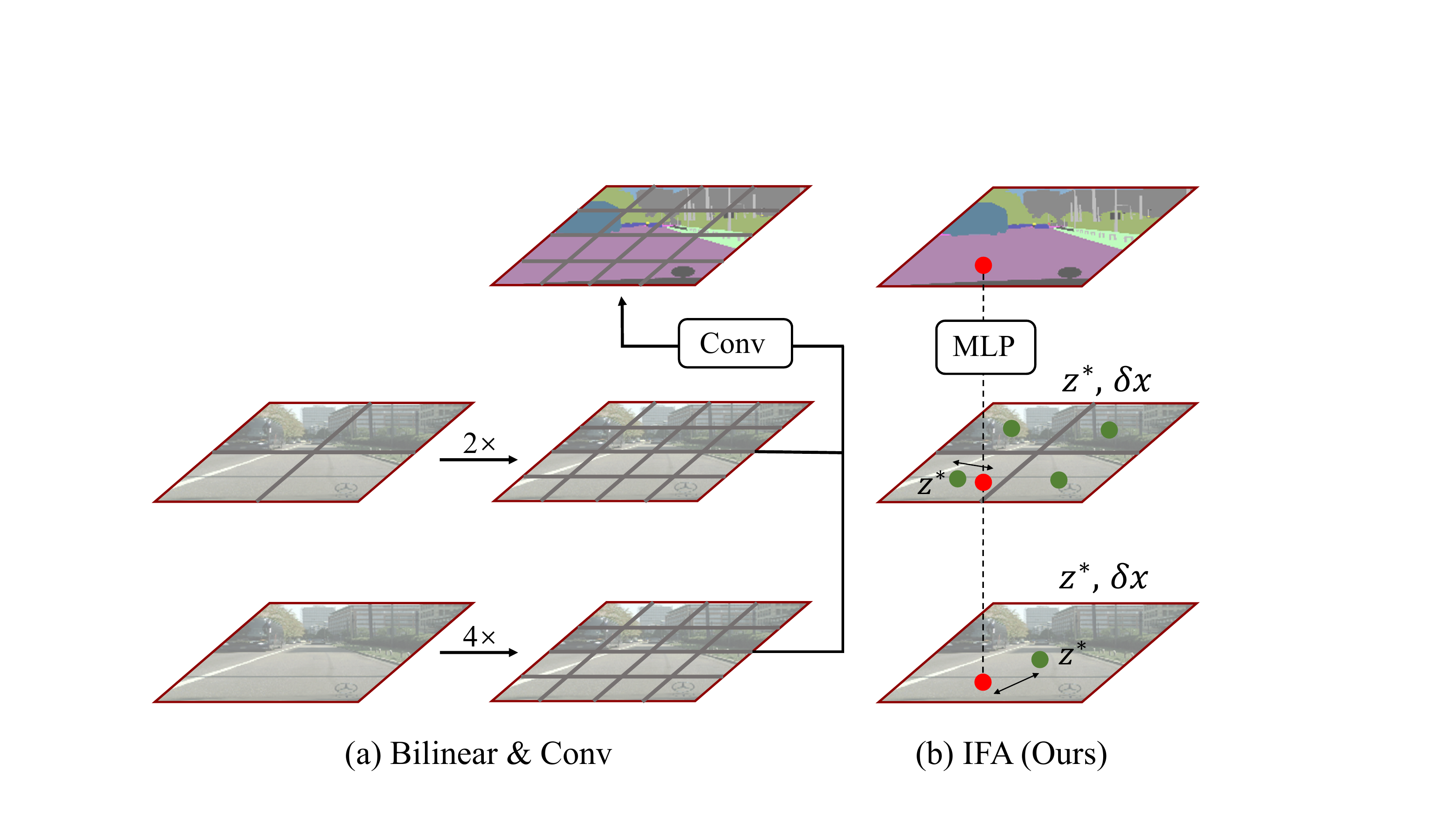}
    \caption{\textbf{Implicit Feature Alignment function (IFA).} (a) Prior works transform feature maps to the same resolution for alignment, where bilinear up-sampling blurs the precise information and convolutions can be inefficient. (b) IFA decodes directly from the original feature maps for arbitrary coordinates. An MLP takes as input the multi-level features around the query coordinate with their relative coordinates and outputs the aggregated information.}
    \label{fig:teaser}
\end{figure}

Most current approaches for semantic segmentation are built upon the Fully Convolutional Network~\cite{long2015fully}. At the heart of these approaches, is to integrate high-level context information with low-level details during segmentation. Empirically, deeper features with coarse resolution correspond to higher semantic information while shallow features in lower layers contain more local details.
To aggregate different levels of information, state-of-the-art approaches such as Feature Pyramid Network (FPN)~\cite{kirillov2019panoptic} and DeepLab V3+~\cite{chen2018encoder} utilize bilinear up-sampling followed by convolutions to align the low resolution deep features with the high resolution shallow features. However, the bilinear up-sampling can blur the precise context learned in deep features and the convolutions are not optimal for speed and memory efficiency, especially for high resolution segmentation where the resolution difference between high-level context and low-level details can be large.

To perform efficient and precise feature alignment, we will require a representation that can flexibly query a location in any resolution and output the corresponding feature values. This formulation corresponds to the implicit neural representation \cite{park2019deepsdf,saito2019pifu,sitzmann2020implicit,jiang2020local} proposed for high-quality 3D shape reconstruction, where a 3D object is represented as a neural network that maps a 3D coordinate to its occupancy in the object~\cite{mescheder2019occupancy} or its signed distance to the object surface~\cite{park2019deepsdf}. This idea is also migrated to the 2D domain~\cite{sitzmann2020implicit,chen2021learning}, where neural functions are proposed as continuous image representations that allow the image to be decoded in an arbitrary resolution. Instead of decoding RGB values, can we use implicit neural representation to perform feature alignment? 

In this paper, we propose a novel Implicit Feature Alignment function (IFA) to efficiently and precisely aggregate features from different levels for semantic segmentation. By forwarding an image to a ConvNet, the features can be viewed as latent codes evenly distributed in spatial dimensions (shown as green dots in Figure~\ref{fig:teaser}). Intuitively, each latent code will represent a field of information. IFA will then decode the output segmentation map at every coordinate independently and parallelly. It takes as inputs the latent codes around the queried coordinate from different levels and the relative coordinates to the latent codes, then outputs the aggregated feature at the queried coordinate for classification, as illustrated in Figure~\ref{fig:teaser}. Take the FPN~\cite{kirillov2019panoptic} model as an example, the original design is to use bilinear up-sampling and convolutions to align multi-layer features. IFA can be a replacement here to align and aggregate the multi-level features. Instead of bilinear up-sampling the features and aligning them in a fixed resolution, IFA allows the features to be learned as precisely representing continuous fields of information. The information in different levels are functions of continuous coordinates, which leads to a precise feature alignment in a resolution-free manner, and we can query the coordinate in arbitrary output resolutions with IFA for semantic segmentation.

We demonstrate the effectiveness of IFA on multiple semantic segmentation architectures and multiple datasets including Cityscapes~\cite{cordts2016cityscapes}, PASCAL Context \cite{mottaghi2014role} and ADE20K~\cite{zhouade}. We replace feature alignment approaches with IFA in different methods, and IFA outperforms the original methods in all cases. IFA shows state-of-the-art computation-accuracy trade-off on all experimented datasets. For performing high resolution image segmentation, we also experiment with reducing the high-level feature map size while maintaining the low-level feature resolution, which improves efficiency and reduces memory cost. IFA has shown a much larger gain on aligning features with larger resolution differences. This not only shows the effectiveness of IFA on precise feature alignment, but also reveals its potential on efficient high resolution semantic segmentation.

To sum up, our contributions are summarized as follows:
\begin{itemize}
    \item We propose a novel implicit neural representation IFA for efficient and precise alignment among multi-level feature maps for semantic segmentation. 

    \item Our IFA can be incorporated with multiple state-of-the-art semantic segmentation models and show improvement in all cases. 

    \item We achieve state-of-the-art computation-accracy trade-off on benchmarks including Cityscapes, PASCAL Context and ADE20K.
\end{itemize}


\section{Related Work}

\noindent\textbf{Semantic Segmentation. }
With the success of deep neural networks~\cite{krizhevsky2012imagenet,simonyan2014very,he2016deep}, semantic segmentation has achieved great progress. Based on Fully Convolutional Network (FCN) \cite{long2015fully}, many works have been proposed. To produce high-resolution semantic segmentation map, spatial and semantic information are both indispensable. There are mainly two lines of research for learning the two kinds of information in semantic segmentation. 

The first stream lies in that the final output of the network contains both spatial and semantic information. Many state-of-the-art methods follow this line to design segmentation head to capture contextual information. From the local perspective, DeepLabV3 \cite{chen2017rethinking} employs multiple atrous convolutions with different dilation rates to capture contextual information, while PSPNet \cite{zhao2017pyramid} utilizes pyramid pooling over sub-regions to harvest information. While from the global perspective, Wang \textit{et al.} \cite{wang2018non} apply the idea of self-attention from transformer \cite{vaswani2017attention} into vision problems and propose the non-local module where correlations between all pixels are calculated to guide the dense contextual information aggregation. Recently, Transformer based models ~\cite{ZhengLZZLWFFXT021,strudel2021segmenter,cheng2021per,xie2021segformer} have achieved great progress in semantic segmentation, while they also suffer from heavy parameters and computation cost.

The other stream dissipates information along outputs of different layers of the network. Hence, the success relies on the feature alignment among the outputs. Our method focuses on this direction.

\noindent\textbf{Feature Alignment. }
A common knowledge in semantic segmentation is that outputs from shallower network layers contain more low-level spatial details, while outputs from deeper network layers possess more high-level semantic information. How to effectively align those features has become a vital problem in this stream of study. Many methods focus on fusing multi-level feature maps for high-resolution spatial details and rich semantics. U-Net \cite{ronneberger2015u}
adds several expanding paths to the contracting path to enable precise localization with the context. Gated-SCNN \cite{takikawa2019gated} proposes a gated mechanism to effectively aggregate low-level details with high-level context. CARAFE ~\cite{WangCX0LL19} introduces a context-aware feature upsampling method, where features inside a predefined region centered at each location are reassembled via a weighted combination.
Semantic FPN \cite{kirillov2019panoptic} applies FPN\cite{lin2017feature} structure to semantic segmentation where multi-level features are aligned by several up-sampling stages consisted of convolution layers and bilinear up-sampling. On top of FPN, SFNet \cite{li2020semantic} proposes the flow alignment module to broadcast high-level context to high-resolution details. AlignSeg \cite{huang2021alignseg} explicitly learns the transformation offsets and adaptively aggregate contextual information for better alignment. However, aforementioned methods for alignment usually take up expensive computations and tend to be inefficient for real-time applications.

\noindent\textbf{Implicit Neural Representation. }
In recent methods in 3D reconstruction, shape, object and scene can be represented by multi-layer perceptron (MLP) that maps coordinates to signals, known as implicit neural representations (INR) \cite{park2019deepsdf,genova2019learning,michalkiewicz2019implicit,MildenhallSTBRN20} since the parameters of the 3D representation are not explicitly encoded by point cloud, mesh or voxel. For example, DeepSDF \cite{park2019deepsdf} learns a set of continuous signed distance functions for shape representation. Later, NeRF \cite{MildenhallSTBRN20} provides a more flexible way for synthesizing novel views of complex scenes. 

Although implicit nerual representation has achieved great progress in 3D tasks, it is relatively under-explored for 2D tasks. \cite{chen2019learning} performs 2D shape generation from latent space for simple digits. \cite{sitzmann2020implicit} replaces ReLU with periodic activation functions inside MLP of implicit neural representation to model natural images in high quality. Recently, LIIF \cite{chen2021learning} applies implicit neural representation to model continuous image representation and UltraSR \cite{xu2021ultrasr} further improves the accuracy by adding spatial encoding for implicit function on 2D images. CRM~\cite{shen2022high} performs image segmentation refinement by using implicit neural representations. 
Implicit PointRend ~\cite{cheng2022pointly} focuses on instance segmentation with point supervision, where implicit function is used to generate different parameters of the point head for each object. Different from these works, we focus on utilizing INR to perform implicit alignment of multi-level features.

\begin{figure*}[t]
    \centering
    \includegraphics[width=0.95\linewidth]{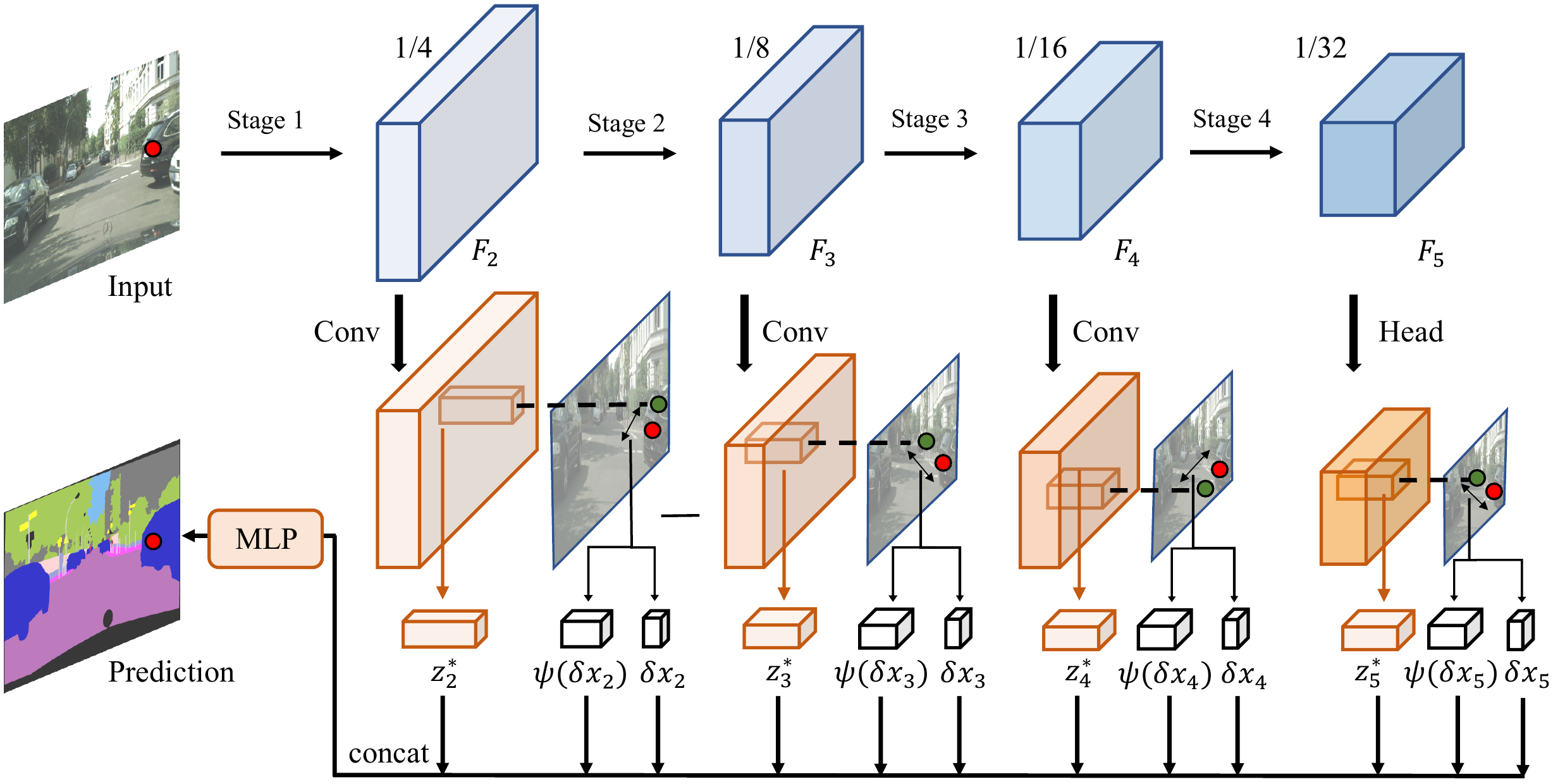}
    \caption{\textbf{Overview of our proposed Implicit Feature Alignment function (IFA).}  The general architecture consists of an encoder part (in blue) and a decoder part (in orange). 
     IFA aligns multi-level feature maps from different stages of the encoder. Each feature map is projected to the same dimension via a convolution layer. We could also project the last feature with a segmentation head like ASPP~\cite{chen2017rethinking}. 
     We view the features in feature maps as latent codes evenly distributed in the 2D space. Given a query coordinate $x_q$, we first find its nearest latent codes $\{z_i^*\}_{i=2}^5$ for each feature map $i$ and use $x^*_i$ to denote the coordinate of $z_i^*$. We then concatenate these latent codes $\{z_i^*\}_{i=2}^5$ and relative coordinates $\{\delta x_i = (x_q - x_i^*)\}_{i=2}^5$, and pass the concatenated vector into an MLP that directly predicts the segmentation label of point $x_q$. 
     The red point refers to the query coordinate $x_q$, while the green point denotes the nearest coordinate $x_i^*$ from $x_q$ on feature $F_i$. 
    }
    \vspace{-1em}
    \label{framework}
\end{figure*}
\section{Method}
In this section, we first introduce the preliminary knowledge about semantic segmentation and reveal the feature alignment within the structures in Section \ref{prelim}. Then we present an overview of the network architectures with the proposed Implicit Feature Alignment Function (IFA) as the alignment method in Section \ref{overall}. Finally, the details of IFA are introduced in Section \ref{IFA}.
\subsection{Preliminary}
\label{prelim}
We will first revisit the basic background of semantic segmentation. Given a RGB image $I \in \mathbb{R}^{3\times H \times W}$, the network aims to produce the segmentation prediction $P \in \mathbb{R}^{N \times H \times W}$, where $H$, $W$ denote the height and width of the input image and $N $ denotes the number of classes. An encoder-decoder paradigm is often adopted, where the encoder applies several down-sampling operations and the decoder employs up-sampling modules to recover the original size. 
To capture rich information, state-of-the-art methods propose to aggregate features from different levels to capture both local details and high-level semantic information.
Following the setting in FPN \cite{kirillov2019panoptic}, different levels of features $\{F_i\}_{i=2}^5$ are extracted from different network stages, where a larger $i$ denotes a deeper stage. $F_i \in \mathbb{R}^{C_i \times H_i \times W_i}$ is a $C_i$ dimensional feature map defined on a spatial grid with size of $H_i\times W_i$ ($H_i=\frac{H}{2^i}$, $W_i=\frac{W}{2^i}$). To compensate the information loss during consecutive down-sampling operations, FPN aims to fuse different levels of feature for better representations. Originally proposed for object detection \cite{lin2017feature}, FPN fuses high-level feature maps with low-level features in a top-down strategy step by step. At each step, FPN fuses high-level feature map with low-level one through several $2\times$ bilinear up-sampling operations with convolution layers.

\subsection{Overall Framework}
\label{overall}
\noindent\textbf{Network Architecture. }
Figure \ref{framework} demonstrates an overview of our network architecture, the general architecture can be described as an encoder part (in blue) and a decoder part (in orange). For a given input image, a bottom-up encoder will encode the image to the feature maps in different levels. A typical example of the encoder is ResNet~\cite{he2016deep}, which generates four feature maps $\{F_i\}_{i=2}^5$ from different stages. For the decoder part, while FPN builds a top-down pathway with bilinear up-sampling, our method instead takes features $\{F_i\}_{i=2}^5$ from the encoder as inputs and decodes the output at every coordinate independently and parallelly, forming a point-independent prediction manner.

\noindent\textbf{Supervision. }Following standard practice in previous state-of-the-art works \cite{zhao2017pyramid,zhang2019acfnet,huclass,li2020semantic}, we add the auxiliary supervision for improving the performance, as well as making the network easier to optimize. Specifically, the output of the third stage of the backbone is further fed into an auxiliary layer to produce an auxiliary prediction, which is supervised with the auxiliary loss. We apply standard cross entropy loss to supervise the auxiliary output and employ OHEM loss \cite{shrivastava2016training} to supervise the main output. 
\subsection{Implicit Feature Alignment Function}
\label{IFA}

In this subsection, we will first introduce an implicit feature function defined on a single feature map. Then we present the details of the position encoding used in our method. Finally, we extend the implicit feature function to the IFA for multi-level features. 

\noindent\textbf{Implicit Feature Function. }One of the main challenges of aggregating information from multi-level feature maps comes from their different resolutions. Up-sampling modules are usually necessary to align them within the same resolution. Our key idea is to define continuous feature maps (i.e. fields of features), which can be decoded at arbitrary coordinates, so that they are aligned in a continuous field and no up-sampling is required.



To define a continuous feature map $M$, we introduce the implicit feature function. It is inspired by recent works of implicit neural representations~\cite{jiang2020local,chen2021learning} for 3D reconstruction and image super-resolution. Implicit feature function defines a decoding function $f_\theta$ (typically an MLP) over a discrete feature map to get the continuous feature map $M$. 
Given the discrete feature map, feature vectors are viewed as latent codes evenly distributed in the 2D space, each of them is assigned with a 2D coordinate. The feature value of $M$ at $x_q$ is defined by
\begin{equation}
    M(x_q) = f_{\theta}(z^*, x_q-x^*),
\end{equation}
where $z^*$ is the nearest latent code from $x_q$ and $x^*$ is the coordinate of latent code $z^*$. To summarize, with the decoding function $f_\theta$, we can define a continuous feature map $M$ over a discrete feature map. In practice, $f_\theta$ is jointly learned with the feature encoder so that the features are learned to precisely represent continuous fields of information.

\noindent\textbf{Position Encoding. }
As discussed in previous works \cite{MildenhallSTBRN20,xu2021ultrasr}, although neural networks can be treated as universal function approximators, the learning power gets limited when directly operated on $xy$ coordinates due to its inferiority at representing high-frequency signals. This is consistent with the discovery of a recent work \cite{rahaman2019spectral} that neural networks are biased towards low-frequency signals and are insensitive to high-frequency signals. Therefore, instead of directly feeding the coordinates to the network, we first encode them with the position encoding function.
Formally, the encoding function we use is:
\begin{equation}
\label{pe}
\begin{aligned}
    \psi(x) = (\text{sin}(\omega_1 x), \text{cos}(\omega_1 x), ..., \\
              \text{sin}(\omega_L x), \text{cos}(\omega_L x)),
\end{aligned}
\end{equation}
where the frequency $\omega_l$ are initialized as $\omega_l = 2e^l, l\in \{1, ..., L\}$ and can be fine-tuned during training, and the encoding function expands the 2D coordinates into the $2L$-dimensional encoding. We also perform experiments on position encoding functions using only $\sin$ or $\cos$ function, and Eq. \ref{pe} performs the best (see Section \ref{ablation} for details). Thus, the final definition of implicit feature function is:
\begin{equation}
\label{if}
    M(x_q) = f_{\theta}(z^*, \psi(x_q - x^*), x_q - x^*),
\end{equation}
where the relative coordinates together with their position encodings are fed into the implicit function.

\noindent\textbf{Feature Alignment. } A direct way to perform feature alignment is to define implicit feature functions and convert each feature map in different levels to a continuous feature map, so that their features can be queried at arbitrary coordinates for alignment. In this subsection, we show that this can be simplified to a more efficient method.


Take aligning the feature maps $\{F_i\}_{i=2}^5$ as an example, we extend the implicit feature function to implicit feature alignment function (IFA), which directly defines a continuous feature map $M$ over multi-level discrete feature maps in different resolutions. Specifically, we define the the value of $M$ at $x_q$ as
\begin{align}
    M(x_q) &= f_\theta\big( \{z^*_i\}_{i=2}^5, \{\psi_i(\delta x_i), \delta x_i\}_{i=2}^5 \big),\\
    \delta x_i &= x_q - x^*_i, \notag
\end{align}
where $i$ denotes the index of feature level, $z^*_i$ is the nearest latent code from $x_q$ at level $i$ and $x^*_i$ is the coordinate of $z^*_i$. We implement $f_\theta$ as concatenating all its input vectors and passing it into an MLP. Intuitively, each latent code still represents a field of feature that can be decoded by relative coordinate, $f_\theta$ can decode the field for each level and model the interaction across different levels at the same time.

The alignment among features produced from different stages is also shown in the lower half of Figure \ref{framework}. For a query coordinate, we obtain the nearest latent codes from each level of features, noted as $\{z^*_i\}_{i=2}^5$, relative coordinates $\delta x_i$ together with the corresponding encoded ones, noted as $\{\psi(\delta x_i)\}_{i=2}^5$. After concatenating the latent codes, relative coordinates and the encoded relative coordinates, we feed them into the MLP of the decoding function $f_\theta$. Given a output resolution, we decode the segmentation map by querying every pixel location independently and parallelly. Therefore, IFA aligns the features in a resolution-free manner and allows decoding to arbitrary resolutions.

Besides FPN, the proposed IFA can be easily applied into other semantic segmentation models that require multi-level feature aggregation, such as DeepLab V3+ \cite{chen2018encoder} and HRNet \cite{wang2020deep}.

\section{Experiments}
\subsection{Datasets and Evaluation Metrics}
\noindent\textbf{Cityscapes. }The Cityscapes dataset~\cite{cordts2016cityscapes} is tasked for urban scene understanding, which contains 30 classes and only 19 classes of them are used for scene parsing evaluation. The dataset contains 5000 finely annotated images and 20000 coarsely annotated images. The size of the images is 2048$\times$1024 pixels. The finely annotated $5,000$ images are split into $2975$, $500$ and $1525$ images for training, validation and testing respectively. We only use finely annotated part in our experiments.

\noindent\textbf{PASCAL Context. }The PASCAL Context is a dataset \cite{mottaghi2014role} is a challenging scene parsing dataset which contains 59 semantic classes and 1 background class. The training set and test set consist of $4,998$ and $5,105$ images respectively.

\noindent\textbf{ADE20K. }The ADE20K dataset \cite{zhouade} is a large scale scene parsing benchmark which contains dense labels of 150 stuff/object categories. The annotated images are divided into 20K, 2K and 3K for training, validation and testing, respectively. 

\noindent\textbf{Evaluation Metric. }The mean of class-wise Intersection over Union (mIoU) is used as the evaluation metric. Number of float-point operations (FLOPs) and number of parameters are also adopted for efficiency evaluations.

\subsection{Implementation Details}
We use ResNet pretrained on ImageNet~\cite{krizhevsky2012imagenet} as our backbone. For Cityscapes dataset, we use stochastic gradient descent (SGD) optimizer with initial learning rate 0.01, weight decay 0.0005 and momentum 0.9. We adopt the `poly' learning rate policy, where the initial learning rate is multiplied by $(1-\frac{\text{iter}}{\text{max}\_\text{iter}})^{0.9}$. We adopt the crop size as $769\times 769$, batch size as 16 and training iterations as 18k. For PASCAL Context dataset, we set the initial learning rate as 0.001, weight decay as 0.0001, crop size as $513\times 513$, batch size as 16 and training iterations as 30K. For ADE20K dataset, we set the initial learning rate as 0.004, weight decay as 0.0001, crop size as $480\times 480$, batch size as 16 and training iterations as 150K.

\begin{table}[t]
\renewcommand\arraystretch{0.9}
\begin{center}
\begin{tabularx}{8.4cm}{p{3.1cm}|X<{\centering}|X<{\centering}|X<{\centering}}
\toprule[1.5pt]
Method  & mIoU(\%) & \#Params & GFLOPs \\
\midrule[1pt]
Bilinear Up-sampling  & 76.52 & 27.7M & 183.4 \\
Nearest Neighbor & 76.32 & 27.7M & 183.4\\
Deconvolution & 72.89 & 29.5M & 304.4 \\
Up-sampling Module & 77.19 & 31.0M & 219.1 \\
CARAFE~\cite{WangCX0LL19} & 76.80 & 29.0M & 190.5\\
AlignSeg~\cite{huang2021alignseg} & 78.50 & 49.7M & 348.6\\
IFA (Ours)  & 78.02 & 27.8M & 186.9\\
\bottomrule[1.5pt]
\end{tabularx}
\end{center}
\vspace{-2mm}
\caption{Performance comparisons of different aligning methods within the FPN structure on Cityscapes \texttt{val} set. GFLOPs calculations adopt $1024\times 1024$ images as input.}
\vspace{-5mm}
\label{am}
\end{table}

\begin{table}[t]
\renewcommand\arraystretch{0.9}
\begin{center}
\begin{tabularx}{6.8cm}{p{4.0cm}|X<{\centering}}
\toprule[1.5pt]
Method  & mIoU(\%) \\
\midrule[1pt]
DeepLab V3+  & 76.69 \\
DeepLab V3+ (IFA)  & \textbf{77.57} \\
\midrule
PSPNet  &73.64 \\
PSPNet (IFA) & \textbf{74.42}  \\
\midrule
HRNet-W18 & 77.60 \\
HRNet-W18 (IFA) & \textbf{78.10} \\
\midrule
HRNet-W48-OCR & 85.80 \\
HRNet-W48-OCR (IFA) & \textbf{86.10}\\
\bottomrule[1.5pt]
\end{tabularx}
\end{center}
\vspace{-2mm}
\caption{Performance of IFA on different segmentation models on Cityscapes \texttt{val} set.}
\vspace{-2em}
\label{others}
\end{table}

\subsection{Results and Ablations}
\label{ablation}
In this subsection, we conduct extensive experiments on the \texttt{val} set of Cityscapes dataset with different settings for our proposed IFA. For all the experiments in this subsection, we use ResNet-50 as the backbone and down-sampling rate as 32 if not specified. All compared methods are evaluated by single-scale inference.

\noindent\textbf{Aligning Method. }
We first compare IFA against commonly used aligning methods, i.e. bilinear up-sampling, nearest up-sampling, deconvolution and up-sampling module (bilinear+convolution), and state-of-the-art methods including CARAFE~\cite{WangCX0LL19} and AlignSeg~\cite{huang2021alignseg}. In particular, we use FPN as the decoder, where the original aligning method is the up-sampling module. We then replace it with other aligning methods. As shown in Table \ref{am}, our proposed IFA performs the best over other baseline methods. While up-sampling module also achieves high performance, its overhead is much higher than the proposed IFA. IFA achieves better results than up-sampling module with 85\% of its computation. 
Moreover, although AlginSeg obtains slightly better results, it takes up almost twice as many parameters and computation cost as ours. Hence, IFA achieves a better trade-off between computational cost and accuracy.

\noindent\textbf{Extension to Other Models. }
Since our proposed IFA targets at aligning features from different levels, we can directly apply it into other segmentation models involving feature alignment such as DeepLab V3+ \cite{chen2018encoder}, PSPNet \cite{zhao2017pyramid} and HRNet \cite{wang2020deep}. In particular, DeepLab V3+ aggregates low-level feature $F_2$ and high-level feature $F_5$ by simple bilinear up-sampling, which can be replaced by IFA. PSPNet aggregate features of different scales produced with different pooling strides by bilinear up-sampling as well. And HRNet also aggregates features of four different scales by bilinear up-sampling. Hence, we plug IFA into these models to replace bilinear up-sampling and perform alignment. The results are presented in Table \ref{others}. IFA improves DeepLab V3+ by 0.9\%, PSPNet by 0.8\%, and HRNet-W18 by 0.5\%. Furthermore, IFA also boost the performance of HRNet-W48-OCR \cite{yuan2020object} by 0.3\%, indicating the strong generalization ability of IFA for different segmentation models.

\begin{table}[t]
\renewcommand\arraystretch{0.9}
\begin{center}
\begin{tabularx}{8.5cm}{p{1.6cm}|p{1.1cm}<{\centering}|p{0.8cm}<{\centering}|p{0.7cm}<{\centering}|X<{\centering}|X<{\centering}}
\toprule[1.5pt]
Method & Stride & Diff & IFA & mIoU(\%) & Gain(\%)\\
\midrule[1pt]
\multirow{6}{*}{\makecell[c]{FPN}}  & \multirow{2}{*}{32} & \multirow{2}{*}{8} &  & 77.19 & \multirow{2}{*}{0.9} \\
 & & & \checkmark & \textbf{78.02} & \\
\cmidrule{2-6}
  & \multirow{2}{*}{64} & \multirow{2}{*}{16} & & 76.52 & \multirow{2}{*}{1.1}\\
 & & & \checkmark & \textbf{77.69} & \\
\cmidrule{2-6}
  & \multirow{2}{*}{128} & \multirow{2}{*}{32} & & 74.88 & \multirow{2}{*}{1.6}\\
 & & & \checkmark & \textbf{76.40} & \\
 \midrule
 \multirow{6}{*}{\makecell[c]{DeepLab\\V3+}} & \multirow{2}{*}{32} & \multirow{2}{*}{8} & & 76.69 & \multirow{2}{*}{0.9}\\
 & & & \checkmark & \textbf{77.57} & \\
\cmidrule{2-6}
 & \multirow{2}{*}{64} & \multirow{2}{*}{16} & & 75.20 & \multirow{2}{*}{1.1}\\
 & & &  \checkmark & \textbf{76.23} &\\
\cmidrule{2-6}
 & \multirow{2}{*}{128} & \multirow{2}{*}{32}& & 70.01 & \multirow{2}{*}{2.1}\\
 & & & \checkmark & \textbf{72.18} & \\
\bottomrule[1.5pt]
\end{tabularx}
\end{center}
\vspace{-2mm}
\caption{Effect of resolution difference on the feature maps of the FPN model. `Stride' denotes the down-sampling rate of the network and `Diff' denotes the scale different between $F_2$ and $F_5$. Results are reported on Cityscapes \texttt{val} set.}
\label{resolution_fpn}
\vspace{-2em}
\end{table}

\noindent\textbf{Resolution Difference. }Commonly in FPN based methods, the largest scale difference between two feature maps is 8 times (between $F_2$ and $F_5$). 
The larger the scale difference is, the harder it is to align feature maps. Moreover, as the improvement of high-resolution image collection tools, we will obtain a large amount of super high resolution data for training and testing. However, with higher-resolution images as input, current methods could be unable to fit into training machines due to limited memory capacity. 

To demonstrate that our proposed IFA can better align feature maps with large scale difference, we alternate the stride of FPN and DeepLab V3+ models and apply IFA. In particular, we add average pooling operation after the first stage of the backbone to further downsample the feature maps, which simultaneously increase both the down-sampling rate of the network and the scale difference between the highest resolution feature ($F_2$) and the lowest resolution feature ($F_5$). The results of experiments on FPN is shown in Table \ref{resolution_fpn}. As the 
scale difference gets larger, the performance gain over the baseline becomes larger as well, 
demonstrating the capability of IFA to align feature map with large scale difference. And the results of experiments on DeepLab V3+ is also shown in Table \ref{resolution_fpn}. Similar conclusion can be obtained. 

\begin{figure}[t]
\begin{minipage}[h]{0.45\textwidth}
\renewcommand\arraystretch{0.9}
\begin{center}
\begin{tabularx}{5cm}{p{2.5cm}|X<{\centering}}
\toprule[1.5pt]
Pos. Enc  & mIoU(\%) \\
\midrule[1pt]
None & 76.88 \\
Coord  & 77.01 \\
Sine & 77.61 \\
Cosine & 77.56 \\
Ours (fixed) & 77.89\\
Ours (learned) & \textbf{78.02}\\
\bottomrule[1.5pt]
\end{tabularx}
\end{center}
\vspace{-6mm}
\captionof{table}{Results for different formations of position encodings on Cityscapes \texttt{val} set.}
\label{pe}
\end{minipage}
\qquad
\begin{minipage}[h]{0.45\textwidth}
\vspace{-4mm}
    \centering
    \includegraphics[width=0.95\linewidth]{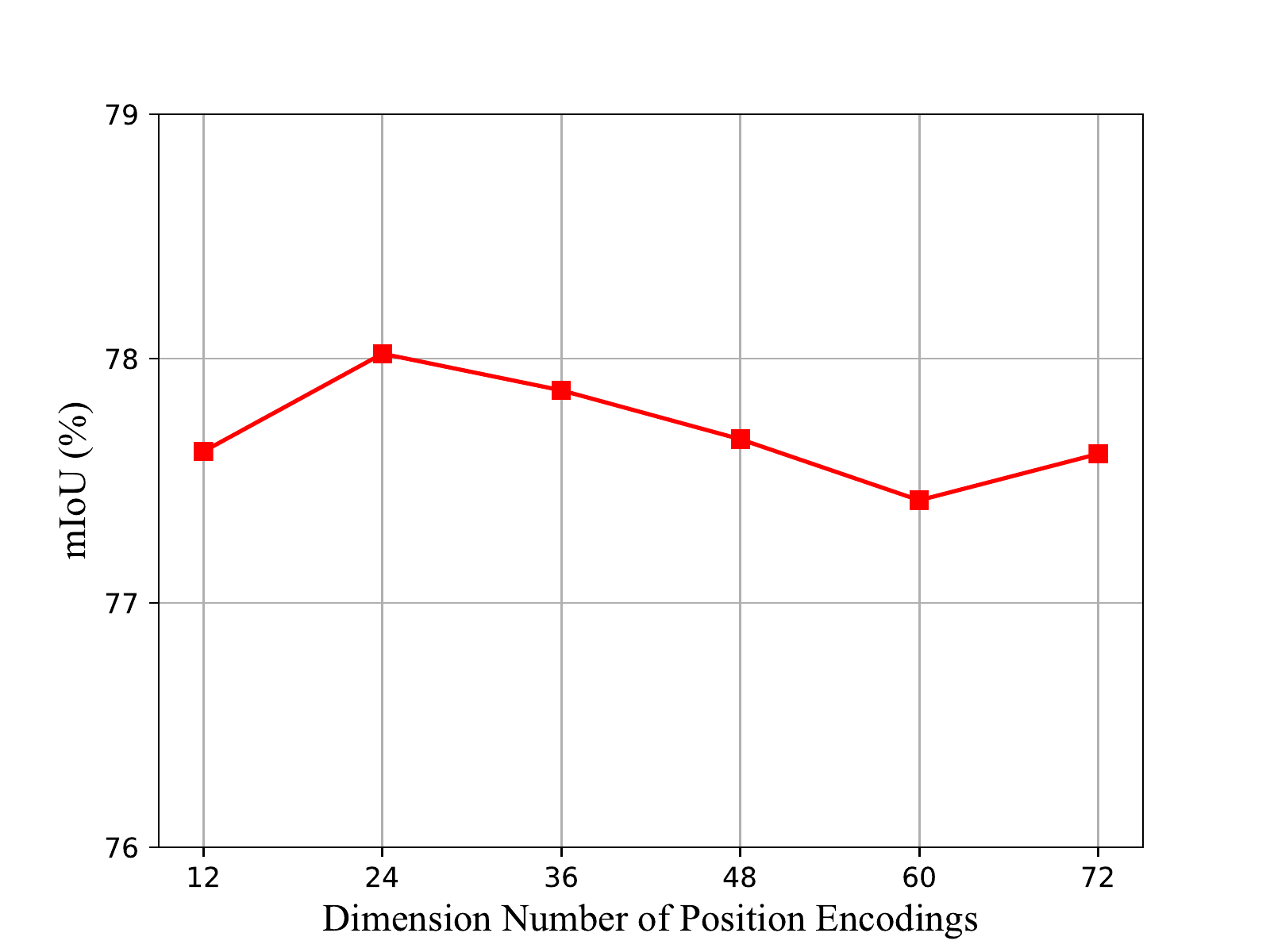}
    \vspace{-2mm}
    \caption{Effect of dimensions of position encoding.}
    \label{pe_dim}
\end{minipage}
\end{figure}
\begin{figure*}[t]
    \centering
    \includegraphics[width=0.9\linewidth]{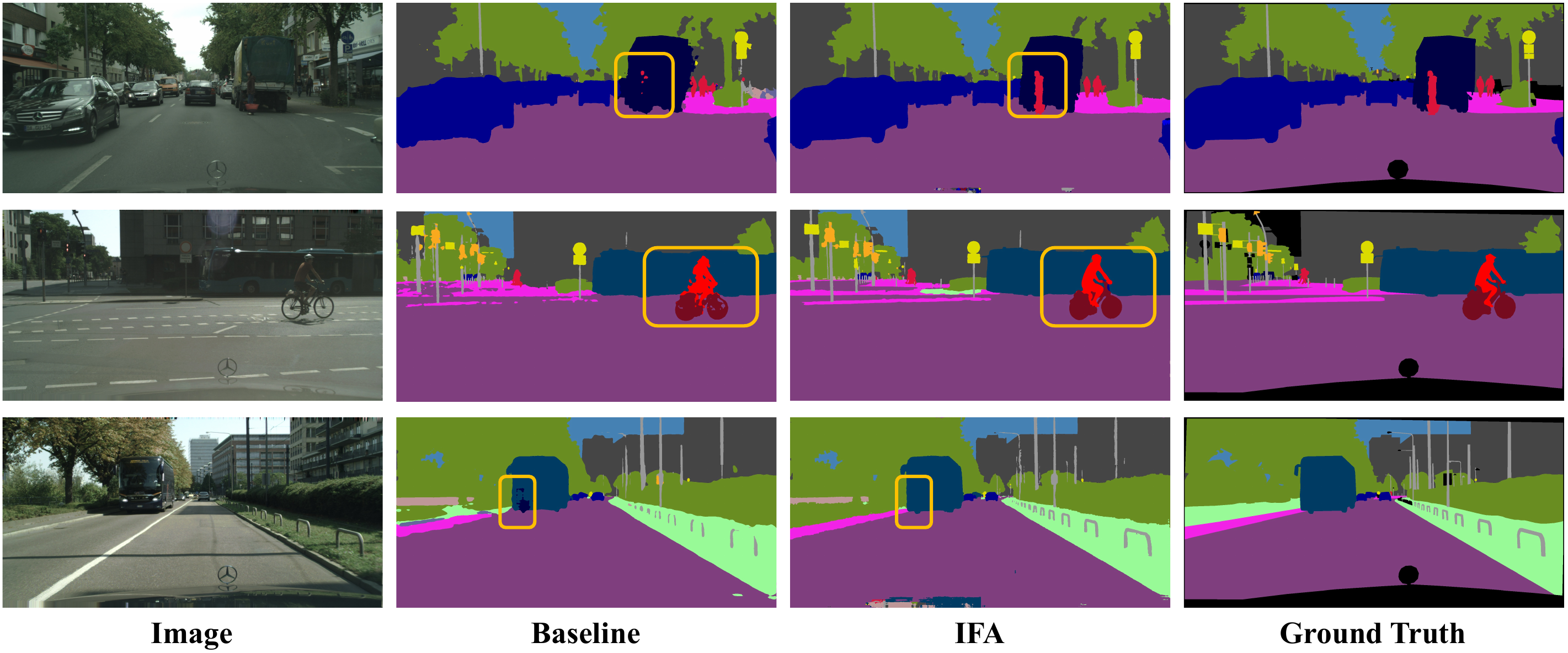}
    \caption{Visualization results on Cityscapes \texttt{val} set. From left to right: input image, predictions made by the FPN baseline, predictions made by the FPN with the proposed IFA and groundtruth map. Yellow squares denote the challenging regions that can be resolved by our proposed IFA.}
    \vspace{-2em}
    \label{visual}
\end{figure*}

\begin{figure}[t]
    \centering
    \includegraphics[width=1\linewidth]{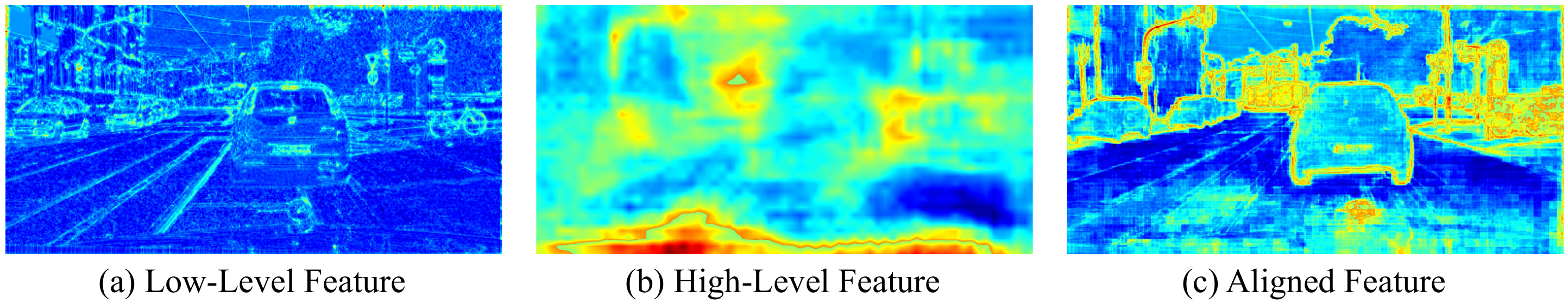}
    \vspace{-4mm}
    \caption{Visualizations of feature maps. (a) Feature map from the first stage of the encoder. (b) Feature map from the last stage of the encoder. (c) Aligned feature from our proposed IFA. }
    \label{vis_feat}
\end{figure}

\noindent\textbf{Position Encoding. }
We further perform experiments to validate the effectiveness of the position encoding inside our proposed Implicit Feature Alignment function (IFA). As illustrated in Table \ref{pe}, we experiment with various formations. 
We first study adding relative coordinates directly, which brings 0.2\% improvement (`Coord'). We also encode the relative coordinates with `Sine' or `Cosine' function, which further improve the results. The learnable frequencies achieves the best performance. The results also demonstrate that position encodings can effectively obtain better results than directly using the spatial coordinates. Moreover, we also investigate the relationship between the dimension number of the position encoding and model's performance. We test a total of six variations: 12, 24, 36, 48, 60 and 72. As shown in Figure \ref{pe_dim}, though the influence of dimension number is not significant, 24 yields the highest performance. Hence, we choose 24 as the dimension number by default.

\begin{table*}[t]
\renewcommand\arraystretch{0.9}
\begin{center}
\begin{tabularx}{12cm}{p{3cm} | p{3cm} | X<{\centering} |X<{\centering}|X<{\centering}}
\toprule[1.5pt]
Method & Backbone & mIoU(\%) & \#Params & GFLOPs \\
\midrule[1pt]
RefineNet~\cite{lin2017refinenet} & Dilated ResNet-101 & 73.6 & - & -\\
GCN~\cite{peng2017large} & Dilated ResNet-101 & 76.9 & - & -\\
SAC~\cite{zhang2017scale} & Dilated ResNet-101 & 78.1& - & - \\
PSPNet~\cite{zhao2017pyramid} & Dilated ResNet-101 & 78.4 & 68.1M & 1104.4\\
DFN~\cite{yu2018learning} & Dilated ResNet-101 & 79.3 & 96.7M & 1185.6 \\
PSANet~\cite{zhao2018psanet} & Dilated ResNet-101 & 80.1 & 89.2M & 1205.7 \\
DenseASPP \cite{yang2018denseaspp} & DenseNet-161 & 80.6 & 39.9M & 640.1 \\
ANNet \cite{zhu2019asymmetric} & Dilated ResNet-101 & 81.3 & 66.5M & 1120.5\\
CCNet~\cite{huang2018ccnet} & Dilated ResNet-101 & 81.4 & 69.8M & 1190.0\\
DANet \cite{fu2019dual} & Dilated ResNet-101 & 81.5 & 69.7M & 1335.9\\
STLNet \cite{zhu2021learning} & Dilated ResNet-101 & \textbf{82.3} & 81.39M &  535.9\\
SETR \cite{ZhengLZZLWFFXT021} & ViT-Large & 81.1 & 318.3M & 2352.0 \\
SegFormer \cite{xie2021segformer} & MiT-B5 & 82.2 &  84.7M & 730.5 \\
\midrule[1pt]
BiSeNet~\cite{yu2018bisenet} & ResNet-18 & 77.7 & 15.6M & 130.2 \\
BiSeNet~\cite{yu2018bisenet} & ResNet-101 & 78.9 & 54.2M & 255.1 \\
SFNet \cite{li2020semantic}& ResNet-18 & 79.5 & 15.9M & 165.4 \\
SFNet \cite{li2020semantic}& ResNet-101 & 81.8 & 55.0M & 459.9 \\
IFA (Ours) & ResNet-18 & 79.3 & 12.3M & 93.3\\
IFA$^*$ (Ours) & ResNet-18 & 79.8 & 16.8M & 98.0\\
IFA (Ours) & ResNet-101 & 81.2 & 46.7M & 262.8\\
IFA$^*$ (Ours) & ResNet-101 & \textbf{82.0} & 64.3M & 281.4 \\
\bottomrule[1.5pt]
\end{tabularx}
\end{center}
\vspace{-2mm}
\caption{Comparisons with state-of-art on the Cityscapes \texttt{test} set. `IFA$^*$' denotes IFA with ASPP module. All methods use multi-scale inference. The GFLOPs is calculated with a image size of $1024\times 1024$. 
}
\vspace{-2.5em}
\label{sota}
\end{table*}

\noindent\textbf{Visualizations of the effect of IFA. }
We further provide comparisons of visualization results on \texttt{val} set of Cityscapes dataset in Figure \ref{visual}. IFA considerably resolves category ambiguities within large objects and produces more precise boundaries of small objects, by effectively aggregating low-level and high-level feature maps. Hence, low-level spatial details and high-level semantic information can be precisely aligned to produce a more accurate prediction.

\noindent\textbf{Visualizations of the feature maps. }
To better demonstrate the effect of IFA, we visualize feature maps from the first stage and final stage of the encoder and the decoded feature from IFA. Bright areas denote the existence of objects.
The visualizations are generated by averaging the features along the channel dimension. 
As shown in Figure \ref{vis_feat}, IFA can effectively leverage spatial details from the low-level feature and semantic information from the high-level feature, thus output a comprehensive feature representation.

\vspace{-0.5em}
\subsection{Comparisons with State-of-the-Arts}
In this subsection, we compare our method with other state-of-the-art methods on three benchmark datasets including Cityscapes, PASCAL Context and ADE20K. Specifically, we choose ResNet-101 as the encoder, FPN as the decoder and replace the up-sampling module in FPN with our proposed IFA. Moreover, to further improve the performance, we add an ASPP \cite{chen2017rethinking} module at the end of the encoder, only performing contextual learning on the last feature map of the pyramid ($F_5$). Similar to other state-of-the-art methods, we also use the multi-scale and flipping strategies for testing to achieve better results. For convenience, we use IFA to represent FPN with IFA.

\begin{table}[t]
\renewcommand\arraystretch{0.9}
\begin{center}
\begin{tabularx}{8.5cm}{p{2.3cm}|p{2.1cm}|X<{\centering}|X<{\centering}}
\toprule[1.5pt]
Method & Backbone & mIoU(\%) & GFLOPs \\
\midrule[1pt]
FCN-8s \cite{long2015fully} & VGG-16 & 37.8 & - \\
DeepLab V2 \cite{chen2017deeplab} & D-ResNet-101 & 45.7 & - \\
RefineNet \cite{lin2017refinenet} & ResNet-152 & 47.3 & - \\
EncNet \cite{zhang2018context} & D-ResNet-101 & 51.7 & - \\
DANet \cite{fu2019dual} & D-ResNet-101 & 52.6 & 296.4\\
ANNet \cite{zhu2019asymmetric} & D-ResNet-101 & 52.8 & 248.3\\
EMANet \cite{Li_2019_ICCV} & D-ResNet-101 & 53.1 & 212.3\\
SETR \cite{ZhengLZZLWFFXT021} & ViT-Large & \textbf{55.8} & 519.5 \\
\midrule[1pt]
SFNet \cite{li2020semantic} & ResNet-101 & 53.8 & 103.0 \\
IFA (Ours) & ResNet-101 & 53.0  & 59.1 \\
IFA$^*$ (Ours) & ResNet-101 & \textbf{53.8} & 63.5\\

\bottomrule[1.5pt]
\end{tabularx}
\end{center}
\vspace{-2mm}
\caption{Comparisons with state-of-art on the PASCAL Context \texttt{test} set. The results are reported under 60 classes (w/o background). `D-' denotes the dilated version of the backbone and `IFA$^*$' denotes IFA with ASPP module. All methods use multi-scale inference. GFLOPs is calculated with a image size of $480\times 480$. }
\vspace{-1.8em}
\label{pascal}
\end{table}

\begin{table}[t]
\renewcommand\arraystretch{0.9}
\begin{center}
\begin{tabularx}{8.5cm}{p{2.3cm}|p{2.1cm}|X<{\centering}|X<{\centering}}
\toprule[1.5pt]
Method & Backbone & mIoU(\%) & GFLOPs \\
\midrule[1pt]
RefineNet~\cite{lin2017refinenet} & ResNet-152 & 40.70 & -\\
PSPNet~\cite{zhao2017pyramid} & D-ResNet-101 & 43.29 & 280.3  \\
CFNet~\cite{zhang2019co} & D-ResNet-101 & 44.89 & -\\
CCNet~\cite{huang2018ccnet} & D-ResNet-101 & 45.22 & 301.2 \\
APCNet~\cite{he2019adaptive} & D-ResNet-101 & 45.38 & - \\
CPNet \cite{yu2020context} & D-ResNet-101 & 46.27 & 314.3 \\
SETR \cite{ZhengLZZLWFFXT021} & ViT-Large & 50.28 & 591.0  \\
SegFormer \cite{xie2021segformer} & MiT-B5 & \textbf{51.80} & 184.6 \\
\midrule[1pt]
SFNet \cite{li2020semantic} & ResNet-101 & 44.67 & 119.4\\
IFA (Ours) & ResNet-101 &  45.23 & 67.1 \\
IFA$^*$ (Ours) & ResNet-101 & \textbf{45.98}  & 72.0\\

\bottomrule[1.5pt]
\end{tabularx}
\end{center}
\vspace{-2mm}
\caption{Comparisons with state-of-art on the ADE20K \texttt{val} set. `D-' denotes the dilated version of the backbone, `IFA$^*$' denotes IFA with ASPP module. All methods use multi-scale inference. GFLOPs is calculated with a image size of $512\times 512$.}
\vspace{-2.5em}
\label{ade}
\end{table}

\noindent\textbf{Cityscapes. }We train the proposed method using both training and validation set of Cityscapes dataset and make the evaluation on the \texttt{test} set by submitting our test results to the official evaluation server. Model parameters and computation FLOPs are also listed for comparison. From Table \ref{sota}, it can be observed that our proposed IFA achieves competitive performance on Cityscapes \texttt{test} set with less computation cost. In particular, our proposed IFA achieves competitive result (81.2) compared with SFNet (81.8) with only 57\% of its computational overhead. And with ASPP module bringing extra 7\% computation, our method (82.0) surpasses SFNet (81.8) while only requiring 61\% of its computation. Moreover, although recent Transformer based model SegFormer achieves slightly better results (82.2) than ours (82.0), it takes extra 32\% parameters and 160\% computation cost. The results demonstrate that our method achieved a better computation-accuracy trade-off compared with other state-of-the-art methods.  
It's also worth noting that SegFormer backbone is pre-trained with stronger data augmentation from the recipe of DeiT~\cite{touvron2021training}, while our backbone is pre-trained with standard data augmentation.

\noindent\textbf{PASCAL Context. }We also conduct experiments on the PASCAL Context dataset. We report the results under 60 classes without the background. Table \ref{pascal} presents the results on the PASCAL Context \texttt{test} set. Our method achieves competitive performance compared with state-of-the-art methods, with less computation cost. In particular, our method surpasses most of the previous methods with much less computation cost and achieves competitive results with SFNet, with only 61\% of its overhead.

\noindent\textbf{ADE20K. }We also carry out experiments on the ADE20K dataset. Performance results on the \texttt{val} set are reported in Table \ref{ade}. Our method achieves competitive result compared with CNN based CPNet, with only 22\% of its computational cost. 

\section{Conclusion}
In this paper, we focus on the feature alignment problem in popular semantic segmentation models involving feature aggregation operations. Hence, we present the Implicit Feature Alignment function (IFA) to perform precise feature alignment among multi-level features. IFA let multi-level features be learned as representing a continuous field of feature and represents the context from different levels as a function of continuous coordinates, which leads to a precise feature alignment in a resolution-free manner. Extensive experiments demonstrate the effectiveness of each component of IFA. Our IFA achieves competitive results on three benchmark datasets, \textit{i.e.,} Cityscapes, PASCAL Context and ADE20K. Importantly, our method obtains a better trade-off between segmentation accuracy and computational cost than previous methods. 



%
%
\bibliographystyle{splncs04}
\bibliography{egbib}
\end{document}